\documentclass[review]{elsarticle}

\usepackage{longtable}
\usepackage{lineno,hyperref}
\usepackage{booktabs}
\usepackage{amssymb}
\usepackage{amsmath}
\usepackage{graphicx}

\usepackage{subcaption}

\DeclareMathOperator*{\argmin}{\arg\!\min}
\modulolinenumbers[5]

\usepackage{comment}
\usepackage{color}

\newcommand\dplus{\mathbin{+\!\!\!+}}
\begin{document}

\begin{frontmatter}

\title{Deep tree-ensembles for multi-output prediction} 

\author[mymainaddress,mysecondaryaddress]{Felipe Kenji Nakano}
\corref{mycorrespondingauthor}
\cortext[mycorrespondingauthor]{Corresponding author}
\ead{felipekenji.nakano@kuleuven.be}

\author[mymainaddress,mysecondaryaddress]{Konstantinos Pliakos}
\ead{konstantinos.pliakos@kuleuven.be}

\author[mymainaddress,mysecondaryaddress]{Celine Vens}
\address[mymainaddress]{KU Leuven, Campus KULAK, Department of Public Health and Primary Care, Etienne Sabbelaan 53, 8500 Kortrijk, Belgium \\}
\address[mysecondaryaddress]{Itec, imec research group at KU Leuven, Etienne Sabbelaan 53, 8500 Kortrijk, Belgium \\}

\ead{celine.vens@kuleuven.be}

\begin{abstract}

Recently, deep neural networks have expanded the state-of-art in various scientific fields and provided solutions to long standing problems across multiple application domains. Nevertheless, they also suffer from weaknesses since their optimal performance depends on massive amounts of training data and the tuning of an extended number of parameters. As a countermeasure, some deep-forest methods have been recently proposed, as efficient and low-scale solutions. Despite that, these approaches simply employ label classification probabilities as induced features and primarily focus on traditional classification and regression tasks, leaving multi-output prediction under-explored. Moreover, recent work has demonstrated that tree-embeddings are highly representative, especially in structured output prediction. In this direction, we propose a novel deep tree-ensemble (DTE) model, where every layer enriches the original feature set with a representation learning component based on tree-embeddings. In this paper, we specifically focus on two structured output prediction tasks, namely multi-label classification and multi-target regression. We conducted experiments using multiple benchmark datasets and the obtained results confirm that our method provides superior results to state-of-the-art methods in both tasks.
\end{abstract}


\end{frontmatter}

\section{Introduction}

Modern technological advances have opened new horizons in many areas of science. Along with new opportunities and possibilities that were deemed unfeasible in the past, this ongoing progress has given birth to new challenges for the scientific community, stimulating new lines of research \cite{Yin2015}. When it comes to the field of machine learning, traditional approaches need to adapt and tackle the new challenges that have emerged (e.g. increased data volume and complexity, need for scalability, etc.) \cite{Jordan2015}. 

In this direction, deep learning has arisen as a cutting edge methodology. In recent years, deep learning has expanded the state-of-the-art in diverse fields of knowledge and application domains, such as medicine \cite{esteva2017}, genomics \cite{eraslan2019}, and self driving cars \cite{badue2020}. 

Apart from its blatant success, deep neural networks may also present some drawbacks. In most of the scenarios, the training consists of tuning a massive number of parameters which demands considerably large training datasets. This often leads to an extreme computational complexity. In addition, deep neural networks often fail to meet high prediction performance when it comes to small scale datasets, which are still mainstream in many application domains. For instance, studies that follow participants over time (e.g. patients in the context of clinical studies) typically have a limited number (often in the range of a few hundreds) of subjects. 

In typical supervised learning, a prediction function is formed on a training set of instances in order to predict a target value \cite{Witten2011}. Each instance is represented by a feature vector and associated with a target, which can be categorical (classification) or numerical (regression). Single target prediction tasks adopt the assumption that an instance is associated with a single output variable. However, this assumption does not hold in many real-world problems, as instances may belong to more than one class at the same time. For example, an article can be associated with multiple topics simultaneously. Multi-output prediction models that learn to predict multiple output variables at the same time are of high importance in this setting. Multi-output prediction, which is also denoted as multi-target prediction \cite{waegeman2019}, is a generalization of multi-target regression and multi-target classification. Multi-label classification can be considered as a special case of multi-target regression (or classification) with only two numerical values (classes) for each target (0 or 1) \cite{tsoumakas2010}. Nowadays, due to the increasing volume and complexity of the stored data, multi-output prediction methods draw growing attention. Such approaches boost prediction performance, exploiting the structure of the target space, while being computationally efficient. 

In this direction, here we focus on multi-output prediction tasks. In particular, we tackle the bottlenecks from deep neural networks by proposing a novel deep tree-ensemble method, where tree-ensembles are built in multiple layers following a deep learning architecture. Our work is motivated by the deep forest approach proposed by Zhou \& Feng \cite{zhou2019deep}, where the model also consists of several layers. In each layer of \cite{zhou2019deep}, a random forest \cite{breiman2001} model outputs a class probability vector which is then used to extend the original input feature set. Different from \cite{zhou2019deep}, in our approach we include a tree-based representation learning step in every layer extending the original input space with low-dimensional tree-embeddings. More specifically, instead of simply using the predictions of the previous layer as extra attributes for the next one, we learn a high dimensional representation based on the decision paths of the trees in the tree-ensemble collection. Next, we project these high dimensional and sparse vectors to a low-dimensional space, yielding compact tree-embeddings which we use to extend the input feature set.

Our main hypothesis is that adding a representation learning component to the deep forest methodology will boost the predictive performance of the method while keeping the computational complexity low. The nodes of each tree of the ensemble are considered clusters that comprise the instances that are associated to them by the recursive partitioning procedure. The ensemble is converted into a binary vector where each item is associated with a cluster node. If a training instance has passed through this node the corresponding item receives the value of 1, otherwise 0. Tree-ensemble models are trained over the original feature set in a supervised learning way and therefore the constructed features incorporate label information. It has been shown that such representations benefit from the learning mechanism of tree-ensembles and are apt to boost the performance of subsequently applied prediction models \cite{vens2011}. Furthermore, we filter and apply weights to this binary representation, based on the number of instances that are contained in a cluster (tree node). The decision paths of the individual trees of the ensemble are transformed to high dimensional and label-aware representations. Finally, we transfer these high dimensional feature vectors to a low dimensional space, removing noise and making the resulting compact tree-embeddings computationally efficient. Low-dimensional tree-embeddings were proved highly representative and effective in multi-output prediction tasks \cite{pliakos2018}.

Driven by the great importance and potential of multi-output prediction as well as by the relative lack of deep-forest methods for multi-output tasks, we deploy our proposed deep tree-ensemble (DTE) method to specifically address the tasks of multi-target regression and multi-label classification. Experiments on benchmark datasets from both domains demonstrate the superiority of our method over state-of-the-art ones.

The main contributions of this work are summarized below:

\begin{itemize}
    \item A new state-of-the-art deep tree-ensemble method for multi-target regression and multi-label classification; 
    
    \item The first study that incorporates representation learning and specifically low-dimensional tree-embeddings into a deep-forest architecture;
    
    \item An effective stopping (pruning) criterion to determine the optimal number of layers as well as mechanisms to surpass overfitting;

    \item An extensive evaluation on 41 datasets, comparing our approach to state-of-the-art methods;
\end{itemize}

The remainder of this paper is organized as follows: Section \ref{section:related} brings a brief overview on multi-target regression, multi-label classification and deep-forest based methods; Section \ref{section:method} introduces our proposed method; Section \ref{section:experimental} presents our experimental setup, and Section \ref{section:results} contains our results and discussion. Finally, Section \ref{section:conclusion} contains our final conclusions and future work directions;

\section{Related work}
\label{section:related}
In this section, we present a brief overview on relevant literature in multi-label classification and multi-target regression, covering the most popular methods in these fields. For a complete literature survey, we refer the reader to \cite{tsoumakas2010,xu2019,borchani2015, zhang2013}.  

Despite the great variety of methods, they can be roughly categorized into two approaches: global and local. The global approach corresponds to models that learn to predict all targets at once. Typically, a traditional algorithm, such as decision trees \cite{blockeel1998}, is adapted to handle the constraints of the problem. Different from that, the local approach builds several models which learn to predict different subsets of outputs \cite{mastelini2020,tsoumakas2010random}. In this case, the predictions of all models are combined to provide the final prediction.  

\subsection{Multi-target regression}

Given a set of instances $X$ and a continuous output set $Y \in R^m$, multi-target regression is defined as the task of learning a function $f$ which, for every unseen instance $x \in X$, outputs $y \in Y$, with $y$ being the target set associated with $x$ \cite{xu2019}.

Kocev \textit{et al.} \cite{kocev2020, kocev2013} proposed global ensembles of predictive clustering trees for structured output tasks. Their experiments showed random forests \cite{breiman2001} as the ensemble method that generally yields superior results. Using deep neural networks, Du and Xu \cite{du2017} proposed a hierarchical neural networks method which employs a divide-and-conquer (local) approach. More specifically, it decomposes the original multi-target problem into several local sub-tasks, and combines their solutions. Also using deep neural networks, Zhen \textit{et al.} \cite{zhen2020} proposed a general model that employs a non-linear layer and a linear low-rank layer to perform direct face alignment via multi-target regression. 
ERC \cite{spyromitros2016} is an ensemble local approach where multiple regressors are chained in a random order. In this method, a separate regressor is built for each target and its predictions are used as extra attributes for the next regressor. Similarly to ERC, DSTARS \cite{mastelini2020} also chains multiple stacked regressors. Its main difference consists of using the feature importance to identify target correlations, in the sense that, targets with positive feature importance values are used as extra attributes for the next regressors.

\subsection{Multi-label classification}

Given a set of instances $X$ and a discrete output set $Y$, multi-label classification is defined as the task of learning a function $f$ which, for every unseen instance $x \in X$, outputs $y \in Y$, with $y$ being the label set associated with $x$ \cite{xu2019}. 

Based on lazy learning, ML-KNN \cite{zhang2007} is an extended version of the traditional k-nearest neighbours, which in addition to finding the closest instances from the training dataset, also employs the concept of maximum a posteriori to predict the labels of new instances. Proposed by Tsoumakas \textit{et al.} \cite{tsoumakas2010random}, Rakel is a local ensemble method which aims to reduce the size of the label set by separating it into smaller random subsets. Next, one model is built per set and the predictions are combined. As an extension of Rakel, Wang \textit{et al.} \cite{wang2019} proposed to select the subsets of labels based on an active learning algorithm, rather than at random. Their results show that the performance can be improved by selecting the best group of labels per subset. Adopting the notion of missing labels, Ma and Chen \cite{ma2020-2} proposed a method that imposes low-rank matrix structures in both global and local manner. More precisely, it enforces local low-rank on predictions with the same labels, whereas the rank is expanded for predictions with different labels.

In the domain of image multi-label classification, several studies have successfully employed convolutional neural networks associated with a label dependence extraction method. For instance, as a recent influential approach, Chen \textit{et al.} \cite{chen2021} proposed a model that exploits graph convolutional networks to build stacked label dependence graphs. Furthermore, Chen \textit{et al.} \cite{chen2019-2} proposed a semantic decoupling method which extracts label specific features and their dependence at the same time. Lastly, You \textit{et al.} \cite{you2020} proposed an adjacency-based method to extract dependence associated to a cross-modality attention mechanism. Despite their wide success, these methods are specifically designed for computer vision tasks, thus we do not consider them further in this work.

\subsection{Deep-Forest based methods}

Similarly to the single output deep forest approach \cite{zhou2019deep}, its multi-label version (MLDF) proposed in \cite{yang2019} employs multiple layers of random forests and extra trees \cite{geurts2006}. However, prediction probabilities are evaluated using measured-based confidence values. More specifically, if the prediction probabilities from the current layer have a better confidence than the ones from the previous layer, they are updated. As an extension of the multi-label deep-forest, Wang \textit{et al.} \cite{wang2020} addressed weak-label learning by using a label complement procedure. At each layer, the label set of the training dataset is complemented using an inner cross-validation scheme, that is, if the label is predicted as positive its label is changed in the training dataset. 

Gao \textit{et al.} \cite{gao2020} proposed a deep-forest variant that includes the oversampling method SMOTE \cite{chawla2002} to address imbalanced tasks. All instances that belong to the minority class and were misclassified in the previous layer are used as input to SMOTE, the new generated instances are concatenated with the dataset and a new layer is built.  

In the oncology field, Su \textit{et al.} \cite{su2019} proposed a deep forest model to predict anti-cancer drug response. The original deep forest was extended by allowing the usage of two feature sets and including a feature optimization method after each layer. Ma \textit{et al.} \cite{ma2020} proposed a cost-sensitive version for price prediction. First, the prices are discretized using K-means, next the authors employed a deep-forest method which penalizes wrong predictions according to how inaccurate they are. In the context of hashing for image retrieval, Zhou \textit{et al.} \cite{zhou2019} demonstrated that the original deep-forest is capable of learning a hashing function that outperforms deep learning hashing methods.

\section{The proposed method}
\label{section:method}

In this section, we thoroughly present our deep tree-ensemble (DTE) method. At first, we introduce the general framework, focusing on its main components. Next, we present a detailed description of the proposed representation learning components. Finally, we introduce the stopping (pruning) criterion that was employed as well as specific characteristics of our architecture.

\subsection{The general framework}

Inspired by recent advances on deep-forest models, our proposed method also employs a cascade of layers of tree ensembles. The main difference relies on the introduced representation learning component at each layer. More specifically, our proposed model extends the original framework by including low-dimensional tree-embeddings, a promising representation learning method, which is further explained in the next subsection.       

It is widely known that diversity among the underlying models is a crucial component in any ensemble method \cite{sagi2018}. Hence, each layer of our model consists of a set of different tree-ensemble models. Let $B = \{\beta_1, \cdots, \beta_i, \cdots, \beta_{|B|} \}$ be the different models. Each $\beta_i$ corresponds to a tree-ensemble model such as random forest (RF) \cite{breiman2001}, extremely randomized trees also known as extra-trees (ET) \cite{geurts2006}, gradient boosting, etc. The number $|B|$ also varies across different applications. Here, we build our deep tree-ensemble (DTE) method using one RF and one ET model ($|B|=2$) per layer. We employ these two models as they are well established and also powerful. Moreover, we follow the setting of predictive clustering trees \cite{blockeel1998} due to its natural clustering and variance reduction characteristics. 

\begin{figure}[ht]
    \centering
    \includegraphics[height=4cm, width = 12cm]{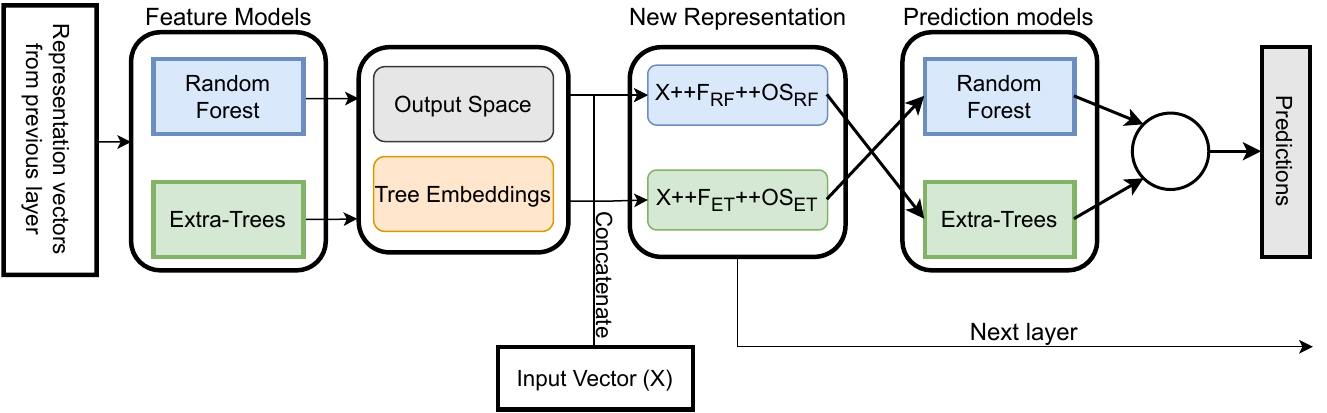}
    \caption{An example of a layer of our proposed model. The box \emph{Output Space} (OS) refers to the prediction features as explained in the text. The arrow \emph{Next layer} links the \emph{New Representations} of this layer to the input box of the next layer, which has the same structure as the illustrated one.}
    \label{figure:deepforest}
\end{figure}

Our proposed framework is illustrated in Figure \ref{figure:deepforest}. Initially, the original feature representation ($X$) is used as input. RF and ET are employed to create their respective tree-embeddings $F$. Let $F_{RF}$ and $F_{ET}$ be the RF and ET based tree-embeddings, respectively.  

Next, the original feature representation ($X$) is concatenated with the generated representations ($F$), and crossed over to be used as input for the models that perform predictions. That is, the $F_{RF}$ is generated in the current layer and given as input to ET, and vice-versa. Our main motivation behind this is that each $\beta_i$ benefits from learning from representations that were yielded by $\beta_j$, where $i \ne j$. This way, we increase the diversity between the predictors. Each model $\beta_i$ is always built using representations that stem from different models. Besides boosting diversity, we assume that this step contributes to avoid overfitting.

Thus, for all layers $L$ with $L \geq 1$, a RF model is trained using $X \dplus F_{ET}$ features, and an ET model with $X \dplus F_{RF}$, where $\dplus$ corresponds to vector concatenation. This can be generalized to $|B|$ tree-ensemble models; $\forall \ \beta_i \in B$ the generated feature set is $X \dplus F_{\beta_1} \dplus F_{\beta_j} \dplus \cdots \dplus F_{\beta_{|B|}}$, where $j \neq i$. These models are chained to the DTE, building a new layer. This procedure is repeated until the stopping criterion is triggered. 

When it comes to final predictions, our method employs all the models $\beta_i \in  B $ of its last layer (w.r.t the stopping criterion). More specifically, it averages the output of all trees in all tree-ensemble models $\beta_i$ to perform predictions.

It is possible to implement different feature representation combinations. For instance, output features as used in \cite{zhou2019deep,yang2019} could be included in our architecture, allowing us to propose four variants of our method:

\begin{itemize}
    \item X\_TE: Original feature representation and \underline{t}ree-\underline{e}mbeddings ($X \dplus F$);
    \item X\_OS: Original features and prediction (i.e., \underline{o}utput \underline{s}pace) features. This variant is identical to the original deep-forest \cite{zhou2019deep}, except for the feature cross-over.
    \item X\_OS\_TE: Original features, tree-embeddings and prediction features;
    \item TE: Only tree-embeddings. A baseline variant which is used to evaluate the importance of concatenating the original feature representation with the generated one.
\end{itemize}

\subsection{Low Dimensional Tree-embeddings}
Originally proposed for biomedical data, tree-embeddings provide more informative and variant representations by exploiting the inner properties of decision trees, in particular, the clustering performed at each split is converted to a new feature representation \cite{pliakos2018}.  

\begin{figure}[ht]
    \centering
    \includegraphics[height = 8cm, width = 12cm]{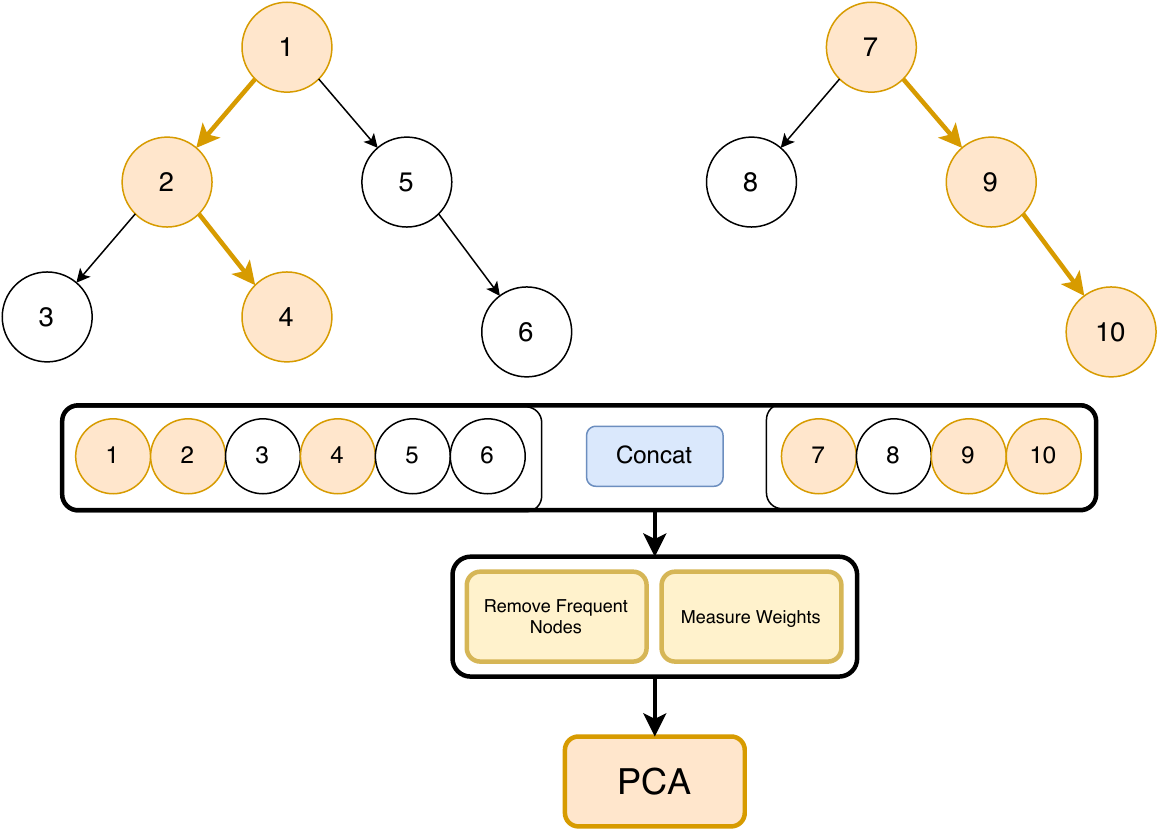}
    \caption{Hypothetical example being classified by leaf nodes 4 and 10.}
    \label{figure:decisionpath}
\end{figure}

Given a trained ensemble of decision trees, tree-embeddings are generated using the following procedure. Initially, all the trees are converted to a binary vector where each position is associated with a node. These vectors are concatenated, resulting in $C = \{c_1, c_2, .... c_{|C|} \}$, $|C|$ being the total number of nodes in the ensemble. By treating $c \in C$ as a feature, we create a new representation $F \in R^{|N| \times |C|}$ where $N$ stands for the number of instances in the dataset. The values in $F$ are instantiated according to how instances traverse the trees. Namely, $F_{ij}$ equals to 1 if such instance $i$ belongs to node (cluster) $j$, otherwise its value is set to 0, as exemplified in Figure \ref{figure:decisionpath}. In more detail, in \ref{figure:decisionpath}, we show a hypothetical instance being classified into leaf nodes 4 and 10. Next, we show the binary vector representation for the instance, where colored circles correspond to vector positions set to 1.

Secondly, we discard the nodes in $F$ (columns $F_{.j}$) which are present in more than $p\%$ of the instances, removing non-meaningful nodes, such as the ones located close to the root and the root itself. Next, the remaining nodes are weighted according to Equation \ref{equation:weights} where nodes that contain many instances are credited a lower weight, and the opposite is valid for more representative nodes. In Equation \ref{equation:weights}, $w_j$ stands for the weight assigned to node $j$, $|c_j|$ is the number of instances that traverse $j$ and $e$ is a small positive value for numerical stability. 

\begin{equation}
\label{equation:weights}
w_j = \frac{1}{log(|c_j|) + e }
\end{equation}

Lastly, the filtered and weighted vectors are transferred from their high dimensional space to a lower dimensional one. Pliakos and Vens \cite{pliakos2018} evaluated several standard methods in this context, such as: PCA, kernel PCA, independent component analysis and truncated singular value decomposition. The results obtained from their experiments have showed that PCA is often associated with superior results. Hence, here we also employ PCA to obtain the final tree-embeddings. Recall that, the input to PCA should be centered before being processed.

This way, through this representation learning component, we yield a compact but yet informative and target-aware representation. By transferring the tree-path features to a low dimensional space we make our approach computationally more efficient removing as well any existing noise. Moreover, by tuning the number of kept components (final dimensionality of the $F$ space) we implicitly assign a latent weight to the added feature set, in relation to the original space $X$. Thus, we can control the number of added features. Recall that, this procedure is performed separately for each ensemble model ($\beta_i$) due to the embeddings cross-over mechanism, as shown in Figure \ref{figure:deepforest}.

Furthermore, considering that tree-embeddings are very representative and label-aware, overfitting can be an issue. Thus, we employ the following procedure: 

\paragraph{Overfitting prevention:
The ensembles in the model itself are not used to generate tree-embedding features. A separate RF and an ET are built using a random sample of the training dataset, these models are used exclusively for the generation of tree-embeddings, and, thus, are not used to perform predictions}

\subsection{Output space features}

Recently, features based on the output of models have been vastly exploited in structured output prediction \cite{mastelini2020,yang2019,read2009}. More precisely, given a chain of models, the outputs (i.e., predictions) of the previous models are used as extra (induced) features for the next one.

These features are known to enrich the representational power by exploiting label correlations among the output set \cite{mastelini2020}. Furthermore, it also allows the models to learn from each other's mistakes \cite{wolpert1992}, enhancing the overall performance. Likewise, output features are commonly exploited in deep-forest models \cite{zhou2019deep,yang2019}.

Thus, we also incorporate them into our model. Moreover, analogously to tree-embeddings, we also employ an overfitting prevention measure to obtain output features.

\paragraph{Overfitting prevention: The ensembles in the model itself are not used to generate output features. Similarly to \cite{zhou2019deep,yang2019}, an inner K-fold cross validation is employed, resulting in 2K separate models (RF + ET). Their averaged outputs, either prediction probabilities or target values, are the output features} \mbox{}\\  

Using hard predictions (0 or 1) as extra attributes, especially for multi-label classification, could also be implemented, as proposed by Read \cite{read2009}. Nonetheless, there has been an increasing tendency to use the raw output of the models due to superior results \cite{read2019}.

\subsection{Pruning criterion}

As for finding the optimal number of layers, different options can be explored. At a first glance, the training could be stopped immediately after any deterioration in the performance. By being more flexible, it is also possible to establish a tolerance level where the training is interrupted if the performance does not improve after a certain number of layers, as proposed by Yang \cite{yang2019}. Both measures may lead to sub-optimal solutions due to premature stops. As a countermeasure, we propose a pruning criterion.

\begin{figure}[ht]
    \centering
    \includegraphics[scale=0.6]{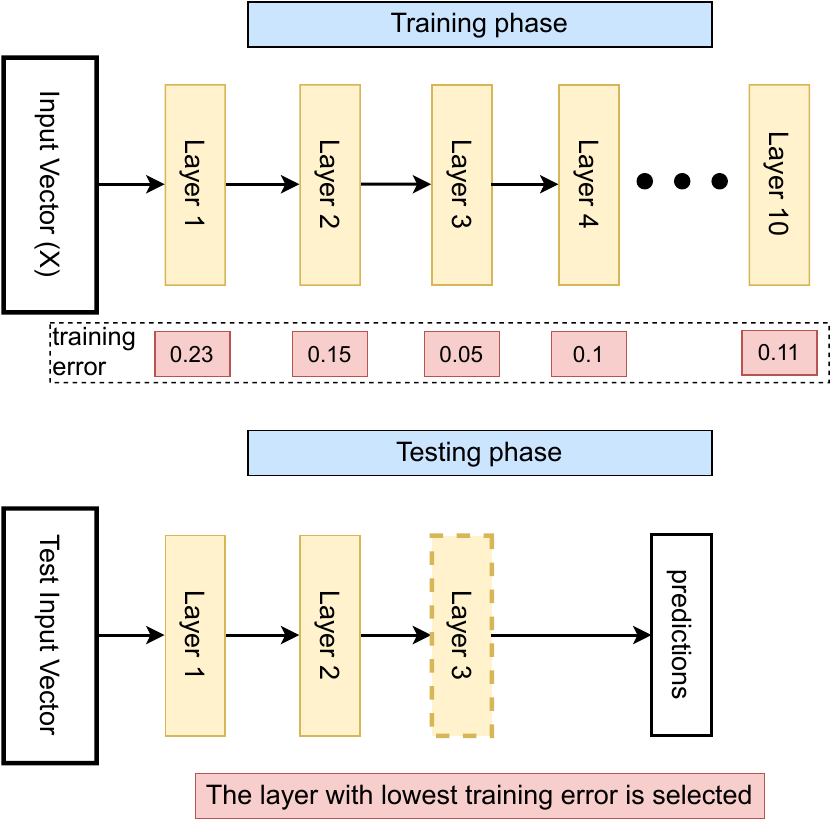}
    \caption{Proposed stopping (pruning) criterion. Select the layer with the lowest training error given a pre-defined number of layers}
    \label{figure:stopcriterion}
\end{figure}

Given a pre-defined number of iterations, we select the layer with the best performance on the training dataset. Thus, all the ensembles trained after the layer with the best performance are pruned. The final model consists of all tree-ensembles up to the layer with the best performance, and the output of this layer yields the final predictions. This pruning criterion is exemplified in Figure \ref{figure:stopcriterion}.

\subsection{Computational complexity}
First, we evaluate the complexity of obtaining tree-embeddings and output space features. Next, we use them to define the complexity of our proposed method. 

As mentioned in the previous section, obtaining tree-embedding features relies mainly on the following steps:

\begin{enumerate}
    \item Building the tree ensembles at each  layer;
    \item Traverse the trees to obtain decision paths;
    \item Apply PCA to obtain the tree-embedding features;
\end{enumerate}

The complexity of measuring tree-embedding features is defined as follows. 1) The cost for building a tree ensemble may vary according to the ensemble itself, hence we generalize this as $Comp_{Model}$; 2) Obtaining decision paths has the same complexity as performing predictions which is $\mathcal{O}(n_{trees}  \times N log(N))$, where N is the number of instances; Finally, 3) PCA is applied to the matrix from the previous step, resulting in $\mathcal{O}(C^2N+C^3)$, $C$  being the number of nodes in the ensemble. Such complexity is associated to PCA because it requires measuring the covariance matrix ($C^2N$) and its eigenvalue decomposition ($C^3$) \cite{golub2013} . Thus, the total complexity, $Comp_{TE}$, is $\mathcal{O} ( Comp_{Model} + n_{trees} \times N log(N) + C^2N+C^3)$.

When it comes to output space features, its complexity depends on:
\begin{enumerate}
    \item Building the tree ensembles at each layer;
    \item Traverse the trees to obtain predictions;    
\end{enumerate}

As previously discussed,  1) the complexity of the first step is $Comp_{Model}$. 2) Likewise, the complexity of performing predictions is defined as $\mathcal{O}(n_{trees} \times N log(N))$. Hence, its total complexity, $Comp_{OS}$, is $\mathcal{O}(Comp_{Model} + (n_{trees} \times N log(N))$.  

The complexity of each variant of our proposed method is defined by combining $Comp_{OS}$ and $Comp_{TE}$. At each layer, it is necessary to compute the features and build the ensembles. Furthermore, a different set of extra features is built for each ensemble. Thus, the complexity of obtaining the features multiplied by the number of models.  Formally, we state the complexity of each of the variants as follows. Here, $N_{models}$ stands for the number of ensembles in each layer. 

\begin{itemize}
\item $Comp_{X\_TE} = \mathcal{O} (N_{models} \times Comp_{Model} + N_{models} \times Comp_{TE}) $;

\item $Comp_{X\_OS} = \mathcal{O} (N_{models} \times Comp_{Model} + N_{models} \times Comp_{OS}) $;

\item $Comp_{X\_TE\_OS} = \mathcal{O} (N_{models} \times Comp_{Model} +  N_{models} \times (Comp_{TE} + Comp_{OS})) $;

\item $Comp_{TE} =   \mathcal{O} (N_{models} \times Comp_{Model} + N_{models} \times Comp_{TE}) $ ;
\end{itemize}
\section{Experimental Setup}
\label{section:experimental}

\subsection{Multi-target regression}
In Section \ref{subsection:datasets}, more specifically, in Table \ref{table:datasetsmtr}, we present the multi-target regression datasets used in this work.

\subsubsection{Datasets}
\label{subsection:datasets}

\begin{table}[htpb]
\tiny
\begin{tabular}{lcccc}
\toprule
Datasets                                   & \#Instances                              & \#Features                              & \#Targets                              & \#Domain                                                                                               \\ \midrule
atp1d                                      & 337                                      & 370                                     & 6                                      & Air ticket price over a day                                                                            \\
atp7d                                      & 296                                      & 370                                     & 6                                      & Air ticket price over 7 days                                                                           \\
oes97                                      & 334                                      & 263                                     & 16                                     & Number of employees per job measured in 1997                                                           \\
oes10                                      & 403                                      & 298                                     & 16                                     & Number of employees per job measured in 2010                                                           \\
rf1    & 9125 & 64  & 8  & Water flow in the Mississippi River                                \\
rf2    & 9125 & 576 & 8  & Water flow in the Mississippi River including forecasting features \\
scm1d  & 9803 & 280 & 16 & Product prices after  1 day                                        \\
scm20d & 8966 & 61  & 16 & Product prices after 20 days                                       \\
edm                                        & 154                                      & 16                                      & 2                                      & Electrical discharge of machinery                                                                      \\
sf1                                        & 323                                      & 10                                      & 3                                      & Occurrence of solar flares over a period of 24h                                                        \\
sf2                                        & 1066                                     & 9                                       & 3                                      & Occurrence of solar flares over a period of 24h(v2)                                                    \\
jura                                       & 359                                      & 15                                      & 3                                      & Heavy metal concentration                                                                              \\
wq                                         & 1060                                     & 16                                      & 14                                     & Fauna and flora in Slovenian rivers                                                                    \\
enb                                        & 768                                      & 8                                       & 2                                      & Heating and cooling of efficient buildings                                                             \\
slump                                      & 103                                      & 7                                       & 3                                      & Concrete quality                                                                                       \\
andro                                      & 49                                       & 30                                      & 6                                      & Water quality                                                                                          \\
osales                                     & 639                                      & 401                                     & 12                                     & Online sales of a product over its first month                                                         \\
scpf                                       & 1137                                     & 23                                      & 3                                      & Number of views, clicks and comments                                                                   \\ 
musicOrigin1                               & 1059                 &68                  &2                  & Geographical origin of music                                                       \\
musicOrigin2                               & 1059                 & 116                 & 2                  & Geographical origin of music                                                       \\
friedman                                   & 500                  & 25                  & 6                  & Data sampled from an uniform  distribution                                         \\
mp5spec                                    & 80                   & 700                 & 3                  & Measurements of corn using a mp5 spectrometer                                      \\
mp6spec                                   & 80                   & 700                 & 3                  & Measurements of corn using a mp6 spectrometer                                      \\
polymer                                    & 41                   & 10                  & 4                  & Polymer processing plants  \\ \bottomrule                                                        
\end{tabular}
\caption{Multi-target regression datasets characteristics}
\label{table:datasetsmtr}

\end{table}

We have collected a total of 24 benchmark datasets from public repositories\footnote{\url{http://mulan.sourceforge.net/datasets-mtr.html}} \footnote{\url{http://people.vcu.edu/~acano/MTR-SVRCC/}} and split them using 10 fold cross-validation. These datasets are available at\footnote{\url{https://itec.kuleuven-kulak.be/?page\_id=8493}}. In terms of characteristics, they range from a very small number of features and targets (enb) to 700 features (mp5spec and mp6spec) and 16 targets (oes97, oes10, scm1d and scm20d).

\subsubsection{Comparison methods}
\begin{itemize}
    \item DSTARS: A recently proposed approach which combines multiple stacked models into a deep structure \cite{mastelini2020}, available at
    \footnote{\url{https://github.com/smastelini/mtr-toolkit}}. 
    \item ERC: An ensemble of regressor chains where one regressor is trained for each target, and outputs are used as extra attributes for the next regressors \cite{spyromitros2016};
    \item MLP: A comparative multi-layer perceptron neural network regressor \footnote{\url{https://scikit-learn.org/stable/}}; 
    \item RF + ET: A shallow version of our proposed method which consists of a single layer with one RF and one ET; \footnote{\url{https://scikit-learn.org/stable/}};
    \item Our proposed method: We compare all the four variants of our method: TE, X\_TE, X\_OS, X\_TE\_OS \footnote{\url{https://itec.kuleuven-kulak.be/?page\_id=8493}}; 
    
\end{itemize}

For RF + ET, we have fixed the number of trees to 150. For the MLP, we have also employed an inner 5-fold cross validation to optimize the number of layers and neurons. More specifically, we have considered the following values: number of neurons \{50,100\} and number of layers \{1,2\}. The activation function and solver were fixed to ReLu and ADAM, respectively.

In its publicly available implementation, DSTARS normalizes all features and targets in the dataset. Consequently, its predictions are also normalized, making them incomparable with the other methods. To avoid this situation, we have removed any normalization procedure involved in DSTARS. We have also set the number of trees to 150. 

As for the ERCs \cite{spyromitros2016}, we have employed regression trees as base models, and 10 randomly ordered chains. In this case, predictions are performed by averaging the output of each model.

For the number of PCA components in our method, we have optimized the number of components considering a percentage of the total number of decision path features\{1 component, 1\% , 5\%, 20\%, 40\%, 60\%, 80\%, 95\%\} multiplied by $min(N,|c|)$, $N$ being the number of instances in the dataset and $|c|$ the number of nodes in the ensemble. The number of trees was fixed to 150 and output features are generated using 5-fold cross validation. As for preventing the overfitting which may be associated with tree-embeddings, we have employed random samples which contain 50\% of the training dataset. Finally, we run our model with a maximum of 10 layers.

\subsubsection{Evaluation Measures}
Similar to related literature, we are employing the aRRMSE (Average Relative Root Mean Squared Error). RRMSE is the square root of the mean squared error (RMSE) of the model divided by the RMSE of a default model that predicts the average target value in the training set. Likewise, aRRMSE is computed by taking the average RRMSE of all targets. Formally, Equation \ref{equation:arrmse} presents aRRMSE where $\widehat{y}$ stands for predicted values, $\overline{y}$ represents the average value, $M$ is total the number of targets and $N$ is the number of testing instances.    

\begin{equation}
\label{equation:arrmse}
    aRRMSE = \frac{1}{M}\sum^{M}_{t=1} \sqrt{\frac{\sum^{N}_{i=1} (y^i_t - \widehat{y}^i_t)^2}{\sum^{N}_{i=1}(y^i_t - \overline{y}^i_t)^2}}
\end{equation}

\subsection{Multi-label classification}
In this section, we provide details on the multi-label experimental setup.

\subsubsection{Datasets}
In Table \ref{table:datasetsmlc}, we present statistical information about the dataset used. 

\begin{table}[htpb]
\tiny
\centering
\begin{tabular}{lccccc}
\toprule
Datasets                                               & \#Instances                                & \#Features                               & \#Labels                                & \# Cardinality                           & \#Application                                                                   \\ \midrule
birds                                                  & 645                                        & 260                                      & 19                                      & 1.01                                     & Species recognition based on chirping                                           \\
CAL500                                                 & 502                                        & 68                                       & 174                                     & 26.04                                    & Instrument, genre and voice identification                                      \\
CHD\_49                                                & 555                                        & 49                                       & 6                                       & 2.58                                     & Coronary heart disease prediction                                                                                \\
emotions                                               & 593                                        & 72                                       & 6                                       & 1.87                                     & Emotion recognition from songs                                                  \\
enron                                                  & 1702                                       & 1001                                     & 53                                      & 3.38                                     & Email classification                                                            \\
flags                                                  & 194                                        & 19                                       & 7                                       & 3.392                                    & Country flag prediction                                                                                \\
genbase            & 662    & 1186 & 27  & 1.25 & Protein function prediction                 \\
Gram\_negative                                         & 1392                                       & 1717                                     & 8                                       & 1.04                                     &        Gram negative bacterial species sub-cellular location                                                                          \\
Gram\_positive                                         &519                                        & 912                                      & 4                                       & 1.04                                     &                                  Gram positive bacterial species sub-cellular location                                               \\
Gram\_positivePseAAC                                   & 519                                        & 912                                      & 4                                       & 1.04                                     &                        Gram positive bacterial species sub-cellular location                                                         \\
LLOG                                                   & 1460                                       & 1004                                     & 75                                      & 1.18                                     &   Text classification                                                                              \\
medical                                                & 978                                        & 1449                                     & 45                                      & 1.25                                     & Diagnosis based on clinical text                                                \\
PlantGO                                                & 978                                        & 3091                                     & 12                                      & 1.07                                     &   Plant species sub-cellular location                                                                              \\
scene                                                  & 2407                                       & 294                                      & 6                                       & 1.07                                     & Image annotation                                                                \\
VirusGO                                                & 207                                        & 749                                      & 6                                       & 1.21                                     &                                             Virus species sub-cellular location                                     \\
VirusPseAAC                                            & 207                                        & 440                                      & 6                                       & 1.21                                     &                     Virus species sub-cellular location                                                           \\
yeast                                                  & 2417                                       & 103                                      & 14                                      & 4.24                                     & Protein function prediction                                                     \\
\bottomrule
\end{tabular}
\caption{Multi-label classification datasets characteristics}
\label{table:datasetsmlc}

\end{table}

As can be seen, the datasets belong to different domains of applications, including sound recognition (birds, emotions and CAL500), image annotation (scene), text classification (enron, LLOG) and medical and biomedical applications such as CHD\_49, gram\_negative,  gram\_positive, gram\_positivePseAAC, plantGO, virusGO, virusPseAAC, medical, genbase and yeast. 

There is also a wide variation in the number of features and labels. Certain datasets present fewer features such as flags and CHD\_49 with 19 and 49, respectively, whereas plantGO presents 1763 features. The number of labels also varies significantly, ranging from 4 up to 174.      

We have collected them from public repositories\footnote{ \url{http://mulan.sourceforge.net/datasets-mlc.html}} \footnote{url{http://www.uco.es/kdis/mllresources/}}, concatenated the train and test subsets, and performed 10 fold cross-validation. These datasets are available at\footnote{\url{https://itec.kuleuven-kulak.be/?page\_id=8493}}.

\subsubsection{Comparison methods}
We have compared the following multi-label methods:

\begin{itemize}
    \item Rakel: An ensemble approach which builds a base classifier per subset of labels \cite{tsoumakas2010random} \footnote{\url{http://scikit.ml/}}. 
    \item MLKNN: A multi-label version of the baseline K-Nearest-Neighbours classifier \cite{zhang2007} \footnote{\url{http://scikit.ml/}};
    \item MLDF: A recently proposed multi-label deep-forest version proposed in \cite{yang2019}, available at \footnote{\url{http://www.lamda.nju.edu.cn/code\_MLDF.ashx}};
    \item MLP: A comparative multi-layer perceptron neural network multi-label classifier \footnote{\url{https://scikit-learn.org/stable/}};    
    \item ECC: An ensemble of classifier chains where one classifier is trained for each class, and outputs are used as extra attributes for the next classifiers \cite{read2009};
    \item RF + ET: A shallow version of our proposed method which consists of a single layer with one RF and one ET; \footnote{\url{https://scikit-learn.org/stable/}};
    \item Our proposed method: We compare all the four variants of our method:  TE, X\_TE, X\_OS, X\_OS\_TE \footnote{\url{https://itec.kuleuven-kulak.be/?page\_id=8493}};  
\end{itemize}
For Rakel, we have employed SVMs with linear kernel as base classifiers, random overlapping subsets of size 3 and 2 $\times$ M classifiers, M being the number of possible labels per dataset. For MLKNN, we have optimized K considering the values \{3,5,7\} and S \{0.5, 0.7, 1.0\} using a 5-fold inner cross validation. 
Similarly to our MTR experiments, we have employed 10 chains following a random order of targets with decision trees as regressors in the ECCs \cite{moyano2017,moyano2018}.

For our main competitor, MLDF, we have employed the same parameters from their original manuscript. That is, one random forest and one extra tree model, 40 trees at the first layer, increasing 20 trees per layer up to 100, 5-fold inner cross-validation to generate the output features. As the stopping criteria, the training is interrupted after the addition of 3 layers without improvement. Since this method is measure dependent, we have compared it using the measures with which it is defined.

The parameters related to the MLP, RF + ET and the variants of our proposed method were optimized in the same manner as in the multi-target regression experiments. In these experiments, however, we employ multiple evaluation measures. We have picked one measure, namely the microAUC, for optimization, meaning that the same model is compared for all evaluation measures. 

As demonstrated in \cite{chen2021,chen2019-2,you2020}, deep convolutional neural networks have been recently applied for image multi-label classification. Despite being very powerful, these approaches are designed for computer vision applications, meaning that raw images are necessary as input. Consequently, they are regarded as out-of-scope. 
\subsubsection{Evaluation Measures}
In Table \ref{table:symbol}, we present the symbols used to describe our multi-label evaluation measures.

\begin{table}[htpb]
\centering
\begin{tabular}{cc}
\toprule
Y                        & True label set                             \\
X                        & Instance set                          \\
Z                        & Predicted labels                      \\
TP                       & True positive                         \\
FP                       & False positive                        \\
FN                       & False negative                        \\
$|S|$                      & Size of set S                         \\
N                        & Number of instances ($|X|$)             \\
M                        & Number of labels in Y ($|Y|$)           \\
$\theta$    & Ranking function                      \\
I                        & Indicator function, returns 1 if its argument evaluates to True \\
$\hat{Y_i}$ & Complement of Y for an instance $i$ \\ 
$\oplus$ & Boolean XOR \\
T &  Threshold values \\
t &  Iterator for threshold values \\
\bottomrule                      
\end{tabular}
\caption{Table with symbols used  to describe the multi-label evaluation measures. }
\label{table:symbol}
\end{table}

The hamming loss (Equation \ref{equation:hamming}) measures the average number of labels wrongly predicted. In this case, we have applied a threshold of 0.5 to obtain $Z$. 

\begin{equation}
\label{equation:hamming}
    Hamming Loss = \sum_i^{N} Z_i \oplus Y_i  
\end{equation}

Based on the ranking of the predicted labels, one error, Equation \ref{equation:oneerror}, measures the average number of times that the label with highest prediction probability (lowest rank) is not a true label.    

\begin{equation}
\label{equation:oneerror}
    One Error = \frac{1}{N}\sum_i^{N} \textit{I }  (\argmin_{y \in Y} \theta(y)   \notin Y_i)  
\end{equation}

Also based on ranking, the ranking error (Equation \ref{equation:ranking}) measures the average proportion of labels that were ranked incorrectly.   
\begin{equation}
    \label{equation:ranking}
    Ranking Error = \frac{1}{N} \frac{1}{|Y_i||\hat{Y_i}|}  |(y_1,y_2): \theta_i(y_1)  > \theta_i(y_2),   (y_1,y_2) \in Y_i \times \hat{Y_i} |
\end{equation}

We are also employing the average precision score \cite{padilla2020}. Due to possible imbalance in the label set, we are employing its micro variant described at Equations \ref{equation:microrecall}, \ref{equation:micro} and \ref{equation:averageprecisionscore}.

\begin{equation}
    \label{equation:microrecall}
    MicroAverageRecall = \sum_j^{M}{\frac{TP_{j}}{TP_{j} + FN_{j}}}     
\end{equation}

\begin{equation}
    \label{equation:micro}
    MicroAveragePrecision = \sum_j^{M}{\frac{TP_{j}}{TP_{j} + FP_{j}}}     
\end{equation} 

\begin{equation}
    \label{equation:averageprecisionscore}
     MicroAveragePrecisionScore =  \sum_t^T \frac{(MicroAverageRecall_t - MicroAverageRecall_{t - 1})}{MicroAveragePrecision_t}
\end{equation}

The microAUC is obtained by measuring the area under the curve generated by using the recall (Equation \ref{eq:recall}) in the x-axis and the false positive rate (FPR, Equation \ref{eq:fpr}) in the y-axis. 
\begin{equation}
\label{eq:recall}
Recall = \frac{TP}{TP + FN}
\end{equation}

\begin{equation}
\label{eq:fpr}
FPR = \frac{FP}{FP + TP}    
\end{equation}

\section{Results and Discussion}
\label{section:results}
Initially, we present the results on multi-target regression, followed by multi-label classification. More results concerning other stopping criteria are available in the Appendix.

\subsection{Multi-target regression}
\begin{table}[htpb]
\centering
\tiny
\begin{tabular}{lcccccccc}
\bottomrule
Datasets $\Downarrow$    & RF+ET                                    & MLP                                      & ERC                                      & X\_TE                                             & TE                              & X\_OS                           & X\_OS\_TE                     & DSTARS                                   \\ \midrule
atp1d        & 0.488                                    & \textbf{0.463} & 0.526                                    & 0.486$^{5.7}$                                     & $0.509^{2.0}$                   & $0.481^{5.6}$                   & $0.479^{7.2}$                 & 0.479                                    \\
atp7d        & 0.814          & 0.957                                    & 1.09                                     & 0.814$^{5.1}$           & \textbf{0.785$^{1.5}$}          & $0.815^{5.8}$                   & $0.804^{4.7}$                 & 1.143                                    \\
oes97        & 0.563                                    & 0.611                                    & 0.648                                    & 0.566$^{5.6}$                                     & 0.625$^{2.8}$                   & 0.575$^{5.3}$                   & $0.575^{5.4}$                 & \textbf{0.508} \\
oes10        & 0.441                                    & 0.512                                    & 0.494                                    & $0.44^{4.7}$                                      & $0.461^{2.7}$                   & $0.447^{6.2}$                   & $0.444^{4.8}$                 & \textbf{0.399} \\
edm          & 0.867                                    & 1.021                                    & 1.096                                    & $0.838^{7.6}$                                     & $0.817^{6.9}$                   & $0.827^{8.8}$                   & $0.819^{8.9}$                 & \textbf{0.799} \\
sf1          & \textbf{0.981} & 1.5                                      & 1.299                                    & $0.997^{7.2}$                                     & $1.006^{7.3}$                   & $1.071^{8.6}$                   & $1.055^{9.0}$                 & 1.111                                    \\
sf2          & 1.402          & 2.766                                    & 2.824                                    & $1.423^{6.0}$                                     & $1.385^{5.2}$                   & $1.49^{7.8}$                    & \textbf{1.356$^{8.0}$}        & 1.615                                    \\
jura         & 0.664                                    & 0.733                                    & 0.685                                    & $0.633^{4.8}$                                     & $0.689^{2.0}$                   & $0.615^{7.0}$                   & $0.637^{7.7}$                 & \textbf{0.56}  \\
wq           & \textbf{0.94}  & 1.165                                    & 0.969                                    & $0.942^{8.3}$                                     & $0.954^{8.2}$                   & $1.001^{9.0}$                   & $0.996^{8.6}$                 & 0.96                                     \\
enb          & 0.178                                    & 0.153          & 0.371                                    & $0.165^{6.4}$                                     & $0.191^{4.3}$                   & \textbf{0.152$^{8.8}$}          & $0.181^{8.7}$                 & 2.529                                    \\
slump        & 0.851                                    & 1                                        & \textbf{0.792} & $0.857^{5.8}$                                     & $0.881^{3.2}$                   & $0.895^{9.0}$                   & $0.89^{8.7}$                  & 0.814                                    \\
andro        & 0.939                                    & 0.909                                    & 1.829                                    & 0.931$^{3.8}$                                     & 1.035$^{3.5}$                   & $0.939^{5.9}$                   & $0.933^{7.3}$                 & \textbf{0.79}  \\
osales       & 0.776                                    & 0.739                                    & 1.158                                    & $0.759^{6.1}$                                     & $0.801^{4.2}$                   & $0.716^{8.2}$                   & $0.712^{8.2}$                 & \textbf{0.653} \\
scpf         & 0.819                                    & 1.17                                     & 1.038                                    & 0.807$^{4.3}$           & \textbf{0.804$^{2.9}$}          & 0.912$^{8.8}$                   & 0.873$^{7.6}$                 & 0.856                                    \\
musicOrigin1 & 0.877                                    & 1.058                                    & 0.875                                    & 0.873$^{8.2}$           & $0.912^{8.4}$                   & \textbf{0.839$^{7.9}$}          & $0.863^{9.0}$                 & 1.493                                    \\
musicOrigin2 & 0.873                                    & 1.045                                    & 0.854                                    & 0.873$^{7.9}$           & $0.916^{8.1}$                   & \textbf{0.851$^{8.8}$}          & $0.873^{8.9}$                 & 1.493                                    \\
friedman     & 0.946                                    & 1.11                                     & 1.02                                     & \textbf{0.936$^{8.0}$}  & $0.95^{7.8}$                    & $1.051^{7.9}$                   & $1.093^{7.3}$                 & 0.936                                    \\
mp5spec      & 1.142                                    & 1.218                                    & \textbf{1.08}  & $1.139^{7.8}$                                     & $1.124^{6.5}$                   & $1.173^{9.0}$                   & $1.216^{9.0}$                 & 2.038                                    \\
mp6spec      & 1.135                                    & 1.185                                    & \textbf{1.054} & $1.133^{7.3}$                                     & $1.118^{4.8}$                   & $1.153^{8.9}$                   & $1.183^{9.0}$                 & 2.036                                    \\
polymer      & 0.722                                    & 0.679                                    & 0.683                                    & 0.678$^{5.4}$           & $0.672^{5.3}$                   & $0.67^{4.4}$                    & \textbf{0.644$^{3.0}$}        & 1.563                                    \\
rf1          & 0.643                                    & 0.656                                    & 0.743                                    & \textbf{0.624$^{2.4}$}  & $0.691^{1.0}$                   & $0.656^{3.0}$                   & $0.644^{3.1}$                 & 1.664                                    \\
rf2          & 0.606                                    & 0.7                                      & 0.635                                    & 0.601$^{3.4}$           & $0.696^{1.0}$                   & \textbf{0.596$^{2.7}$}          & $0.619^{3.7}$                 & 1.537                                    \\
scm1d        & 0.536                                    & 0.498                                    & \textbf{0.429} & $0.554^{3.4}$                                     & $0.641^{1.0}$                   & $0.536^{4.0}$                   & $0.548^{4.0}$                 & 0.471                                    \\
scm20d       & \textbf{0.748} & 0.786                                    & 0.777                                    & 0.766$^{3.6}$                                     & $0.81^{1.0}$                    & $0.76^{4.0}$                    & $0.76^{4.0}$                  & 0.749                                    \\ \midrule
mean         & 0.789                                    & 0.943                                    & 0.958                                    & \textbf{0.785$^{5.78}$} & 0.811$^{4.23}$ & 0.801$^{6.89}$ & 0.8$^{6.91}$ & 1.133                                    \\                           \bottomrule
\end{tabular}
\caption{Performance in terms of aRRMSE for multi-target regression. The numbers above the measures correspond to the mean number of layers in the final model.}
\label{table:mtrresults}
\end{table}

\begin{figure}[ht]
    \centering
    \includegraphics[scale=0.7]{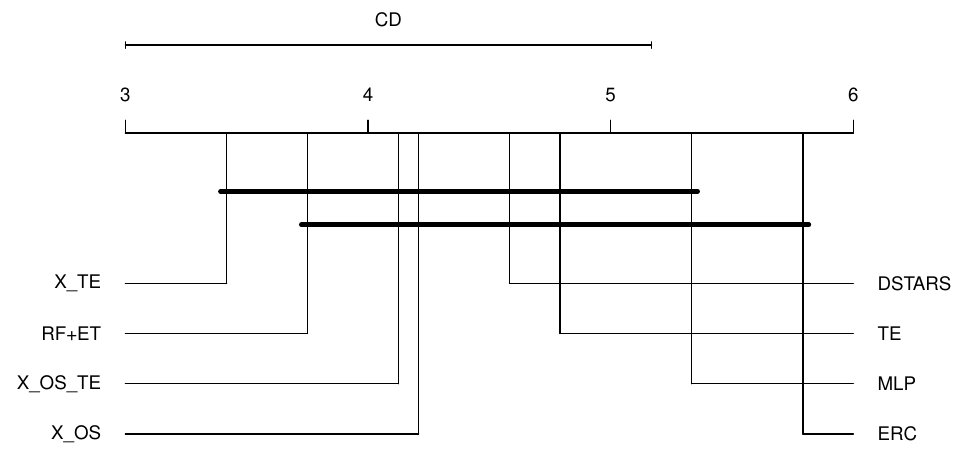}
    \caption{Friedman-Nemenyi comparing the multi-target regression methods. }
    \label{figure:mtrnemenyi}
\end{figure}

As can be seen in Table \ref{table:mtrresults}, our proposed X\_TE provides the best results on average. When compared to the main competitor (DSTARS), X\_TE was ranked in a higher position according to the Friedman-Nemenyi test presented in Figure \ref{figure:mtrnemenyi}.

The second best method was the baseline ensemble RF + ET. In few cases, it managed to be the best method, nonetheless,  on average, it is slightly worse than our X\_TE variant. These results are very surprising, since RF + ET is regarded as a relatively simple approach. 
Despite being very similar between themselves, when compared to each other, X\_TE provided the best performance in more datasets than RF + ET. In 15 out of 24 datasets, X\_TE was the superior method, whereas RF + ET yielded the best performance in 7 datasets and tied in 2 cases. These results are reinforced by the Friedman-Nemenyi diagram shown in Figure \ref{figure:mtrnemenyi}, where X\_TE is ranked higher than RF + ET.  
This finding also validates our choice of using RF + ET as building blocks for more complex models. Despite its relative simplicity, it managed to be superior than complex models such as the state-of-art comparison methods. 

The performance of the comparison methods was unexpected. Even though ERC managed to achieve the highest performance in some cases, on average it is considerably worse than X\_TE. Likewise, DSTARS also had the upper-hand in several cases, nonetheless when all datasets are considered, its performance is underwhelming. This is due to their oscillating performance where they perform better in some cases, but very poorly in others. A similar behaviour was also noticed for the MLP. 

When comparing our variants with each other, the TE variant provided inferior results in many cases. It also limited itself to shallower models due to its performance sequentially decreasing on the training dataset. This is an indication that tree-embeddings should be used to augment the original features and not entirely replace them. Hence, it highlights the necessity of using the original features (X) with the extra features.

Moreover, the variants that employ output space features, X\_OS and X\_OS\_TE, were also associated to the best performance in a few datasets. Despite being almost equivalent among themselves, they are not competitive when compared to X\_TE. Such evidence attests our hypothesis that learning a representation component at each layer of our proposed method (DTE) leads to superior predictive performance.

These results also validate our stopping (pruning) criterion. In comparison to the results obtained with different stopping criteria, our performance is superior to immediately stopping the training after deterioration (Appendix Table 1), and it approaches the best performance in the test (Appendix Table 2).       

\subsection{Multi-label classification}
\begin{table}[htpb]
\tiny
\begin{tabular}{lcccccccccc}
\toprule
Datasets $\Downarrow$          & RF+ET          & ML-KNN         & Rakel          & MLP            & ECC            & MLDF                            & X\_TE                                   & TE                              & X\_OS                          & X\_OS\_TE                       \\ \midrule
birds              & 0.047          & 0.061          & 0.053          & 0.067          & 0.055          & $0.044^{6.9}$                   & $0.047^{4.78}$                          & $0.05^{2.89}$                   & $0.042^{4.56}$                 & \textbf{0.041$^{5.33}$}         \\
CAL500             & 0.138          & 0.161          & \textbf{0.138} & 0.139          & 0.16           & $0.227^{1.1}$                   & \textbf{0.138$^{1.44}$}                 & \textbf{0.138$^{1.56}$}         & $0.144^{2.33}$                 & $0.142^{1.67}$                  \\
CHD\_49            & \textbf{0.286} & 0.327          & 0.287          & 0.306          & 0.338          & $0.314^{3.0}$                   & $0.29^{6.0}$                            & $0.29^{5.67}$                   & $0.312^{9.0}$                  & $0.311^{9.0}$                   \\
emotions           & \textbf{0.182} & 0.27           & 0.25           & 0.216          & 0.243          & $0.198^{4.4}$                   & $0.189^{6.22}$                          & $0.192^{5.67}$                  & $0.192^{8.78}$                 & $0.191^{8.89}$                  \\
enron              & 0.047          & 0.06           & 0.051          & 0.055          & 0.056          & $0.063^{2.1}$                   & $0.047^{4.78}$                          & $0.048^{7.22}$                  & \textbf{0.046$^{9.0}$}         & \textbf{0.046$^{8.44}$}         \\
flags              & \textbf{0.248} & 0.315          & 0.321          & 0.297          & 0.301          & $0.278^{3.9}$                   & $0.24^{6.67}$                           & $0.251^{6.67}$                  & $0.285^{9.0}$                  & $0.27^{8.78}$                   \\
genbase            & 0.047          & \textbf{0.001} & 0.047          & \textbf{0.001} & \textbf{0.001} & $0.025^{5.8}$                   & $0.017^{4.67}$                          & $0.005^{4.44}$                  & $0.042^{4.89}$                 & $0.029^{6.0}$                   \\
Gram\_negative     & 0.046          & 0.021          & 0.076          & 0.014          & \textbf{0.012} & $0.017^{7.9}$                   & $0.018^{5.56}$                          & $0.016^{3.89}$                  & $0.033^{5.89}$                 & $0.017^{7.67}$                  \\
Gram\_positive     & 0.067          & 0.04           & 0.149          & 0.038          & \textbf{0.033} & $0.048^{4.7}$                   & $0.035^{5.0}$                           & \textbf{0.033$^{7.0}$}          & $0.036^{5.89}$                 & $0.034^{6.33}$                  \\
GrampositivePseAAC & 0.239          & 0.27           & 0.269          & 0.274          & 0.303          & $0.239^{4.5}$                   & \textbf{0.233$^{5.78}$}                 & $0.249^{4.11}$                  & $0.251^{7.11}$                 & $0.248^{7.78}$                  \\
LLOG               & 0.153          & \textbf{0.108} & 0.16           & 0.11           & 0.124          & $0.154^{3.4}$                   & $0.123^{3.22}$                          & $0.124^{1.44}$                  & $0.146^{6.22}$                 & $0.155^{6.33}$                  \\
medical            & 0.022          & 0.017          & 0.028          & \textbf{0.012} & 0.01           & $0.014^{10.7}$                  & $0.016^{6.33}$                          & $0.017^{1.0}$                   & $0.017^{6.56}$                 & $0.015^{7.11}$                  \\
PlantGO            & 0.062          & 0.054          & 0.09           & 0.045          & 0.046          & $0.045^{10.4}$                  & $0.051^{5.22}$                          & \textbf{0.043$^{2.67}$}         & $0.056^{4.89}$                 & $0.046^{7.78}$                  \\
scene              & 0.123          & 0.115          & 0.155          & 0.105          & 0.151          & $0.114^{7.3}$                   & $0.119^{2.22}$                          & $0.123^{2.11}$                  & $0.095^{8.78}$                 & \textbf{0.091$^{8.89}$}         \\
VirusGO            & 0.119          & 0.066          & 0.246          & 0.054          & \textbf{0.035} & $0.054^{5.3}$                   & $0.077^{3.44}$                          & $0.069^{2.78}$                  & $0.095^{5.67}$                 & $0.071^{4.0}$                   \\
VirusPseAAC        & 0.21           & 0.231          & 0.24           & 0.242          & 0.288          & $0.226^{2.5}$                   & \textbf{0.206$^{5.56}$}                 & $0.22^{7.0}$                    & $0.222^{4.22}$                 & $0.215^{5.67}$                  \\
yeast              & 0.195          & 0.205          & 0.221          & 0.199          & 0.219          & $0.204^{7.0}$                   & \textbf{0.194$^{8.44}$}                 & $0.202^{6.22}$                  & $0.204^{8.89}$                 & $0.2^{9.0}$                     \\ \midrule
mean               & 0.131          & 0.137          & 0.164          & 0.128          & 0.14           & 0.133$^{5.35}$ & \textbf{0.12$^{5.02}$} & 0.122$^{4.26}$ & 0.13$^{6.57}$ & 0.125$^{6.98}$ \\ \bottomrule
\end{tabular}
\caption{Hamming Distance using MicroAUC to prune}
\label{table:hamming}
\end{table}

\begin{table}[htpb]
\tiny
\begin{tabular}{lcccccccccc}
\toprule
Datasets    $\Downarrow$       & RF+ET          & ML-KNN & Rakel & MLP   & ECC   & MLDF                            & X\_TE                                    & TE                             & X\_OS                           & X\_OS\_TE                       \\ \midrule
birds              & 0.068          & 0.473  & 0.537 & 0.427 & 0.274 & $0.329^{4.6}$                   & $0.073^{4.78}$                           & $0.094^{2.89}$                 & \textbf{0.062$^{4.56}$}         & $0.07^{5.33}$                   \\
CAL500             & \textbf{0.175} & 0.707  & 0.782 & 0.762 & 0.33  & $0.643^{1.5}$                   & \textbf{0.175$^{1.44}$}                  & $0.177^{1.56}$                 & $0.196^{2.33}$                  & $0.191^{1.67}$                  \\
CHD\_49            & 0.203          & 0.534  & 0.449 & 0.485 & 0.316 & $0.508^{4.5}$                   & $0.203^{6.0}$                            & \textbf{0.201$^{5.67}$}        & $0.224^{9.0}$                   & $0.24^{9.0}$                    \\
emotions           & \textbf{0.145} & 0.576  & 0.52  & 0.538 & 0.268 & $0.405^{4.6}$                   & $0.149^{6.22}$                           & $0.154^{5.67}$                 & $0.159^{8.78}$                  & $0.171^{8.89}$                  \\
enron              & \textbf{0.076} & 0.553  & 0.734 & 0.488 & 0.258 & $0.311^{5.9}$                   & \textbf{0.076$^{4.78}$}                  & $0.088^{7.22}$                 & $0.081^{9.0}$                   & $0.085^{8.44}$                  \\
flags              & \textbf{0.185} & 0.51   & 0.484 & 0.499 & 0.324 & $0.606^{2.5}$                   & \textbf{0.185$^{6.67}$}                  & $0.201^{6.67}$                 & $0.233^{9.0}$                   & $0.241^{8.78}$                  \\
genbase            & 0.034          & 0.017  & 1.0   & 0.011 & 0.004 & $0.273^{5.5}$                   & \textbf{0.007$^{4.67}$}                  & \textbf{0.007$^{4.44}$}        & $0.016^{4.89}$                  & $0.009^{6.0}$                   \\
Gram\_negative     & 0.017          & 0.081  & 0.482 & 0.057 & 0.033 & $0.222^{9.1}$                   & \textbf{0.01$^{5.56}$}                   & $0.013^{3.89}$                 & $0.011^{5.89}$                  & $0.009^{7.67}$                  \\
Gram\_positive     & 0.051          & 0.088  & 0.297 & 0.086 & 0.049 & $0.531^{6.9}$                   & $0.036^{5.0}$                            & \textbf{0.035$^{7.0}$}         & $0.042^{5.89}$                  & $0.032^{6.33}$                  \\
GrampositivePseAAC & 0.26           & 0.696  & 0.537 & 0.611 & 0.448 & $0.805^{2.4}$                   & $0.25^{5.78}$                            & \textbf{0.249$^{4.11}$}        & $0.259^{7.11}$                  & $0.263^{7.78}$                  \\
LLOG               & 0.157          & 0.527  & 0.701 & 0.521 & 0.314 & $0.69^{1.0}$                    & \textbf{0.131$^{3.22}$}                  & $0.134^{1.44}$                 & $0.152^{6.22}$                  & $0.191^{6.33}$                  \\
medical            & 0.035          & 0.344  & 1.0   & 0.312 & 0.108 & $0.185^{11.1}$                  & $0.03^{6.33}$                            & $0.036^{1.0}$                  & $0.033^{6.56}$                  & \textbf{0.028$^{7.11}$}         \\
PlantGO            & 0.041          & 0.347  & 1.0   & 0.316 & 0.149 & $0.132^{10.4}$                  & $0.036^{5.22}$                           & $0.045^{2.67}$                 & $0.034^{4.89}$                  & \textbf{0.03$^{7.78}$}          \\
scene              & 0.113          & 0.382  & 0.664 & 0.314 & 0.298 & $0.475^{12.2}$                  & 0.109$^{2.22}$                  & $0.13^{2.11}$                  & \textbf{0.085}$^{8.78}$                  & $0.098^{8.89}$                  \\
VirusGO            & 0.088          & 0.168  & 0.796 & 0.152 & 0.071 & $0.193^{5.6}$                   & $0.039^{3.44}$                           & $0.052^{2.78}$                 & $0.066^{5.67}$                  & \textbf{0.034$^{4.0}$}          \\
VirusPseAAC        & \textbf{0.222}          & 0.7    & 0.806 & 0.732 & 0.496 & $0.511^{3.9}$                   & \textbf{0.222$^{5.56}$}                  & $0.241^{7.0}$                  & $0.233^{4.22}$                  & $0.261^{5.67}$                  \\
yeast              & 0.167          & 0.451  & 0.577 & 0.44  & 0.238 & $0.462^{7.5}$                   & \textbf{0.166}$^{8.44}$                           & $0.176^{6.22}$                 & $0.187^{8.89}$                  & $0.195^{9.0}$                   \\ \midrule
mean               & 0.12           & 0.421  & 0.669 & 0.397 & 0.234 & 0.428$^{5.84}$ & \textbf{0.112$^{5.02}$} & 0.12$^{4.26}$ & 0.122$^{6.57}$ & 0.126$^{6.98}$ \\ \bottomrule
\end{tabular}
\caption{Label ranking using MicroAUC to prune}
\label{table:labelranking}

\end{table}

\begin{table}[htpb]
\tiny
\begin{tabular}{lcccccccccc}
\toprule
Datasets    $\Downarrow$       & RF+ET        & ML-KNN         & Rakel          & MLP            & ECC            & MLDF                            & X\_TE                           & TE                              & X\_OS                          & X\_OS\_TE                       \\ \midrule
birds              & 0.509        & 0.533          & 0.537          & 0.545          & 0.549          & $0.549^{7.0}$                   & $0.509^{4.78}$                  & $0.532^{2.89}$                  & $0.468^{4.56}$                 & \textbf{0.463$^{5.33}$}         \\
CAL500             & \textbf{1.0} & \textbf{1.0}   & \textbf{1.0}   & \textbf{1.0}   & \textbf{1.0}   & \textbf{1.0$^{1.0}$}            & \textbf{1.0$^{1.44}$}           & \textbf{1.0$^{1.56}$}           & $1.0^{2.33}$                   & \textbf{1.0$^{1.67}$}           \\
CHD\_49            & 0.836        & 0.886          & \textbf{0.804}          & 0.848          & 0.89           & $0.885^{3.2}$                   & $0.822^{6.0}$                   & $0.834^{5.67}$                  & 0.812$^{9.0}$         & $0.814^{9.0}$                   \\
emotions           & 0.689        & 0.82           & 0.77           & 0.77           & 0.792          & $0.727^{3.7}$                   & $0.682^{6.22}$                  & $0.695^{5.67}$                  & 0.661$^{8.78}$        & \textbf{0.654$^{8.89}$}                  \\
enron              & 0.898        & 0.917          & 0.989          & 0.879          & 0.884          & $0.916^{11.3}$                  & $0.899^{4.78}$                  & $0.905^{7.22}$                  & $0.879^{9.0}$                  & \textbf{0.866$^{8.44}$}         \\
flags              & 0.84         & 0.932          & 0.861          & 0.915          & 0.828          & $0.86^{2.7}$                    & \textbf{0.754$^{6.67}$}         & $0.823^{6.67}$                  & $0.766^{9.0}$                  & \textbf{0.754$^{8.78}$}         \\
genbase            & 1.0          & 0.029          & 1.0            & 0.024          & \textbf{0.022} & $0.404^{4.8}$                   & $0.31^{4.67}$                   & $0.099^{4.44}$                  & $0.931^{4.89}$                 & $0.713^{6.0}$                   \\
Gram\_negative     & 0.326        & 0.113          & 0.489          & 0.081          & \textbf{0.07}           & $0.198^{8.4}$                   & $0.121^{5.56}$                  & $0.096^{3.89}$                  & $0.23^{5.89}$                  & 0.113$^{7.67}$         \\
Gram\_positive     & 0.256        & 0.096          & 0.301          & 0.1            & \textbf{0.09}           & $0.102^{5.7}$                   & $0.113^{5.0}$                   & $0.096^{7.0}$                   & $0.111^{5.89}$                 & 0.096$^{6.33}$         \\
GrampositivePseAAC & 0.87         & 0.699          & \textbf{0.541} & 0.652          & 0.761          & $0.919^{2.7}$                   & $0.806^{5.78}$                  & $0.763^{4.11}$                  & $0.598^{7.11}$                 & $0.592^{7.78}$                  \\
LLOG               & 1.0          & \textbf{0.859} & 1.0            & \textbf{0.859} & \textbf{0.859} & $1.0^{2.8}$                     & \textbf{0.859$^{3.22}$}         & \textbf{0.859$^{1.44}$}         & $0.948^{6.22}$                 & $0.979^{6.33}$                  \\
medical            & 0.804        & 0.513          & 1.0            & 0.411          & \textbf{0.322} & $0.45^{15.0}$                   & $0.569^{6.33}$                  & $0.614^{1.0}$                   & $0.627^{6.56}$                 & $0.512^{7.11}$                  \\
PlantGO            & 0.658        & 0.416          & 1.0            & 0.367          & \textbf{0.323} & $0.421^{16.5}$                  & $0.526^{5.22}$                  & $0.393^{2.67}$                  & $0.556^{4.89}$                 & $0.443^{7.78}$                  \\
scene              & 0.619        & 0.456          & 0.69           & 0.413          & 0.605          & $0.491^{9.0}$                   & $0.58^{2.22}$                   & $0.594^{2.11}$                  & $0.368^{8.78}$                 & \textbf{0.333$^{8.89}$}         \\
VirusGO            & 0.567        & 0.251 & 0.862          & 0.23           & \textbf{0.134}          & $0.258^{5.3}$                   & $0.337^{3.44}$                  & $0.278^{2.78}$                  & $0.443^{5.67}$                 & $0.299^{4.0}$                   \\
VirusPseAAC        & 0.952        & \textbf{0.791} & 0.856          & 0.871          & 0.909          & $0.976^{3.1}$                   & $0.92^{5.56}$                   & $0.867^{7.0}$                   & $0.883^{4.22}$                 & $0.856^{5.67}$                  \\
yeast              & 0.855        & 0.799          & 0.97           & 0.814          & 0.858          & $0.831^{7.1}$                   & $0.828^{8.44}$                  & $0.84^{6.22}$                   & $0.769^{8.89}$                 & \textbf{0.761$^{9.0}$}          \\ \midrule
mean               & 0.746        & 0.595          & 0.804          & \textbf{0.575} & 0.582          & 0.646$^{6.43}$ & 0.626$^{5.02}$ & 0.605$^{4.26}$ & 0.65$^{6.57}$ & 0.603$^{6.98}$ \\ \bottomrule
\end{tabular}
\caption{One error using MicroAUC to prune}
\label{table:oneerror}

\end{table}

\begin{table}[htpb]
\tiny
\begin{tabular}{lccccccccc}
\toprule
Datasets    $\Uparrow$       & RF+ET          & ML-KNN & Rakel & MLP   & ECC   & X\_TE                                   & TE                              & X\_OS                           & X\_OS\_TE                       \\ \midrule
birds              & 0.599          & 0.09   & 0.053 & 0.12  & 0.315 & $0.595^{4.78}$                          & $0.502^{2.89}$                  & \textbf{0.628$^{4.56}$}         & $0.618^{5.33}$                  \\
CAL500             & \textbf{0.485} & 0.244  & 0.253 & 0.259 & 0.334 & $0.484^{1.44}$                          & $0.477^{1.56}$                  & $0.453^{2.33}$                  & $0.461^{1.67}$                  \\
CHD\_49            & \textbf{0.707}          & 0.549  & 0.589 & 0.57  & 0.621 & $0.706^{6.0}$                           & $0.697^{5.67}$                  & 0.669$^{9.0}$          & $0.649^{9.0}$                   \\
emotions           & 0.751          & 0.433  & 0.466 & 0.514 & 0.601 & $0.752^{6.22}$                          & $0.739^{5.67}$                  & \textbf{0.754$^{8.78}$}         & $0.718^{8.89}$                  \\
enron              & 0.615          & 0.271  & 0.269 & 0.32  & 0.434 & \textbf{0.618$^{4.78}$}                 & $0.588^{7.22}$                  & $0.598^{9.0}$                   & $0.591^{8.44}$                  \\
flags              & \textbf{0.83}  & 0.624  & 0.621 & 0.638 & 0.737 & $0.826^{6.67}$                          & $0.791^{6.67}$                  & $0.782^{9.0}$                   & $0.769^{8.78}$                  \\
genbase            & 0.686          & 0.977  & 0.047 & 0.977 & 0.988 & $0.979^{4.67}$                          & \textbf{0.982$^{4.44}$}         & $0.886^{4.89}$                  & $0.974^{6.0}$                   \\
Gram\_negative     & 0.954          & 0.86   & 0.501 & 0.901 & 0.937 & \textbf{0.981$^{5.56}$}                 & $0.976^{3.89}$                  & $0.966^{5.89}$                  & $0.98^{7.67}$                   \\
Gram\_positive     & 0.958          & 0.87   & 0.574 & 0.875 & 0.94  & $0.979^{5.0}$                           & $0.976^{7.0}$                   & $0.97^{5.89}$                   & \textbf{0.981$^{6.33}$}         \\
GrampositivePseAAC & 0.513          & 0.317  & 0.354 & 0.335 & 0.338 & \textbf{0.516$^{5.78}$}                 & $0.495^{4.11}$                  & $0.499^{7.11}$                  & $0.492^{7.78}$                  \\
LLOG               & 0.656          & 0.528  & 0.308 & 0.521 & 0.625 & \textbf{0.7$^{3.22}$}                   & $0.686^{1.44}$                  & $0.605^{6.22}$                  & $0.542^{6.33}$                  \\
medical            & 0.718          & 0.468  & 0.028 & 0.588 & 0.77  & \textbf{0.796$^{6.33}$}                 & $0.784^{1.0}$                   & $0.745^{6.56}$                  & $0.776^{7.11}$                  \\
PlantGO            & 0.795          & 0.503  & 0.09  & 0.562 & 0.68  & $0.834^{5.22}$                          & \textbf{0.846$^{2.67}$}         & $0.813^{4.89}$                  & $0.844^{7.78}$                  \\
scene              & 0.711          & 0.514  & 0.37  & 0.561 & 0.523 & $0.719^{2.22}$                          & $0.689^{2.11}$                  & \textbf{0.746$^{8.78}$}         & $0.714^{8.89}$                  \\
VirusGO            & 0.804          & 0.734  & 0.231 & 0.779 & 0.89  & \textbf{0.929$^{3.44}$}                 & $0.918^{2.78}$                  & $0.853^{5.67}$                  & $0.927^{4.0}$                   \\
VirusPseAAC        & 0.435          & 0.268  & 0.236 & 0.261 & 0.279 & \textbf{0.439$^{5.56}$}                 & $0.431^{7.0}$                   & $0.425^{4.22}$                  & $0.417^{5.67}$                  \\
yeast              & 0.724          & 0.532  & 0.49  & 0.543 & 0.638 & \textbf{0.727$^{8.44}$}                 & $0.705^{6.22}$                  & $0.702^{8.89}$                  & $0.684^{9.0}$                   \\ \midrule
mean               & 0.702          & 0.517  & 0.322 & 0.548 & 0.626 & \textbf{0.74$^{5.02}$} & 0.722$^{4.26}$ & 0.711$^{6.57}$ & 0.714$^{6.98}$ \\ \bottomrule
\end{tabular}
\caption{Micro average precision score using MicroAUC to prune}
\label{table:microaverage}

\end{table}

\begin{table}[htpb]
\tiny
\begin{tabular}{lcccccccc}
\toprule
Datasets  $\Uparrow$         & RF+ET          & Rakel & MLP   & ECC   & X\_TE                                    & TE                              & X\_OS                           & X\_OS\_TE                       \\  \midrule
birds              & 0.925          & 0.5   & 0.601 & 0.744 & $0.921^{4.78}$                           & $0.879^{2.89}$                  & \textbf{0.931$^{4.56}$}         & $0.923^{5.33}$                  \\
CAL500             & \textbf{0.823} & 0.597 & 0.605 & 0.741 & \textbf{0.823$^{1.44}$}                  & $0.82^{1.56}$                   & $0.8^{2.33}$                    & $0.805^{1.67}$                  \\
CHD\_49            & \textbf{0.794} & 0.711 & 0.688 & 0.722 & $0.793^{6.0}$                            & $0.788^{5.67}$                  & $0.759^{9.0}$                   & $0.74^{9.0}$                    \\
emotions           & \textbf{0.876} & 0.688 & 0.71  & 0.789 & $0.875^{6.22}$                           & $0.864^{5.67}$                  & $0.872^{8.78}$                  & $0.858^{8.89}$                  \\
enron              & \textbf{0.923} & 0.644 & 0.734 & 0.839 & \textbf{0.923$^{4.78}$}                  & $0.909^{7.22}$                  & $0.919^{9.0}$                   & $0.914^{8.44}$                  \\
flags              & \textbf{0.84}  & 0.678 & 0.702 & 0.774 & \textbf{0.84$^{6.67}$}                   & $0.812^{6.67}$                  & $0.799^{9.0}$                   & $0.791^{8.78}$                  \\
genbase            & 0.962          & 0.5   & 0.991 & 0.997 & $0.989^{4.67}$                           & \textbf{0.992$^{4.44}$}         & $0.98^{4.89}$                   & $0.986^{6.0}$                   \\
Gram\_negative     & 0.99           & 0.745 & 0.964 & 0.979 & \textbf{0.995$^{5.56}$}                  & $0.993^{3.89}$                  & $0.993^{5.89}$                  & \textbf{0.995$^{7.67}$}         \\
Gram\_positive     & 0.973          & 0.802 & 0.946 & 0.975 & $0.989^{5.0}$                            & $0.989^{7.0}$                   & $0.983^{5.89}$                  & \textbf{0.992$^{6.33}$}         \\
GrampositivePseAAC & 0.762          & 0.642 & 0.62  & 0.631 & \textbf{0.77$^{5.78}$}                   & $0.76^{4.11}$                   & $0.755^{7.11}$                  & $0.754^{7.78}$                  \\
LLOG               & 0.855          & 0.6   & 0.758 & 0.815 & \textbf{0.881$^{3.22}$}                  & $0.877^{1.44}$                  & $0.836^{6.22}$                  & $0.77^{6.33}$                   \\
medical            & 0.965          & 0.5   & 0.834 & 0.941 & $0.971^{6.33}$                           & $0.966^{1.0}$                   & $0.97^{6.56}$                   & \textbf{0.972$^{7.11}$}         \\
PlantGO            & 0.968          & 0.5   & 0.827 & 0.914 & $0.974^{5.22}$                           & $0.972^{2.67}$                  & $0.974^{4.89}$                  & \textbf{0.978$^{7.78}$}         \\
scene              & 0.907          & 0.646 & 0.817 & 0.801 & \textbf{0.909$^{2.22}$}                  & $0.89^{2.11}$                   & $0.92^{8.78}$                   & $0.894^{8.89}$                  \\
VirusGO            & 0.926          & 0.555 & 0.904 & 0.963 & \textbf{0.977$^{3.44}$}                  & $0.973^{2.78}$                  & $0.946^{5.67}$                  & \textbf{0.977$^{4.0}$}          \\
VirusPseAAC        & \textbf{0.774} & 0.552 & 0.584 & 0.628 & \textbf{0.774$^{5.56}$}                  & 0.762$^{7.0}$                   & $0.757^{4.22}$                  & 0.743$^{5.67}$                  \\
yeast              & 0.847          & 0.686 & 0.745 & 0.807 & \textbf{0.848$^{8.44}$}                  & $0.835^{6.22}$                  & $0.827^{8.89}$                  & $0.815^{9.0}$                   \\  \midrule
mean               & 0.889          & 0.62  & 0.766 & 0.827 & \textbf{0.897$^{5.02}$} & 0.887$^{4.26}$ & 0.884$^{6.57}$ & 0.877$^{6.98}$ \\ \bottomrule
\end{tabular}
\caption{Micro AUC}
\label{table:microauc}

\end{table}

As detailed in Tables \ref{table:hamming}, \ref{table:labelranking}, \ref{table:oneerror}, \ref{table:microaverage} and \ref{table:microauc}, our variants are often associated to the best performances in most of the measures. Similar to the MTR experiments, the X\_TE performs better in the majority of the experiments as stated by the hamming loss, label ranking, micro average precision score and microAUC.

According to both hamming and ranking losses, X\_TE surpasses the competitors in a large number of datasets, leading to a better performance overall. The superiority of our method is more noticeable according to the micro average precision score, as shown in Table \ref{table:microaverage}. In this case, X\_TE yields a considerably higher mean micro average precision score.

A similar behaviour was perceived for microAUC where X\_TE normally provides the best results. Moreover, we could notice that the inclusion of tree-embeddings leads to an improvement in almost all cases when compared to the original representation (RF + ET).

When analyzing one error, the neural network, MLP, was superior on average, followed by another competitor method, ECC. These results are unexpected since these methods failed to achieve a competitive performance according to the other measures. In particular, these methods were exceptionally better than X\_TE in some datasets, such as genbase, medical and plantGO, nonetheless our variant X\_OS\_TE provides better results in more datasets to the point that it is ranked higher according to its Frieman-Nemenyi test (Appendix Figure 1).

\begin{figure}[ht]
    \includegraphics[scale=0.6]{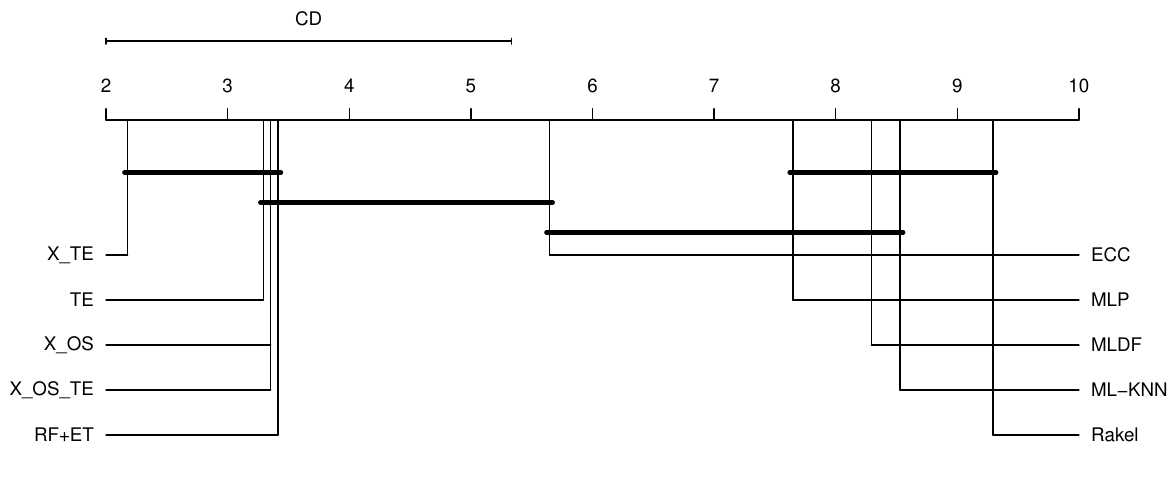}
    \caption{Friedman-Nemenyi test considering the Label Ranking error.}
    \centering
    \label{figure:rankingnemenyi}
\end{figure}

Our main competitor, MLDF, struggled to achieve a competitive performance. With the exception of a few cases, its performance was underwhelming even when compared to the baselines for all three available evaluation measures \footnote{The extra features employed by MLDF are measure specific, and they are not defined for micro average precision score and microAUC.}. This finding is rather surprising at first since it is fairly similar to our variant X\_OS. However, there are some major differences between the two methods. First, there is a different optimization mechanism for the number of trees. Second, the extra features used at each layer are different, since X\_OS directly concatenates the output features, and MLDF employs a measure-aware feature reuse mechanism. Third, our method also employs cross-over between the extra feature sets to prevent overfitting. The difference is highlighted in Figure \ref{figure:rankingnemenyi} where MLDF is associated with a low rank, making it statistically significantly different from all variants of our method. 

Moreover, the TE variant struggles again to perform competitively against X\_TE, instigating the necessity of incorporating the original features. Likewise, the stopping criterion adopted by our method approximates the best possible performance (Appendix Table 4).

When analyzing the CAL500 dataset, we have noticed that the X\_TE variant is constantly superior than the variants X\_OS and X\_OS\_TE in all evaluation measures. We believe that this is related to the number of labels in its label set. Since CAL500 has a relatively large label set, its output features are often similar to sparse vectors, meaning that, in this case, they may hinder the models performance. This is further exemplified by the performance provided by the baseline (RF + ET) which is always superior to all variants with output features (X\_OS and X\_OS\_TE) in this dataset. When compared to each other, X\_TE is superior to RF + ET. This motivates the usage of tree-embeddings in other applications with larger label sets. Furthermore, at this point, it can be deduced that methods which use directly the output variables to augment the input feature space, such as MLDF, could not be applied to extreme multi-label data. They would lead to extremely high dimensional feature sets, dramatically increasing their computational demands.

In some datasets, such as gram\_negative,  gram\_positive and grampositivePseAAC, the performance of most of evaluated methods was close to flawless. Differently from that, unexpected results were obtained in the CAL500 dataset considering the one error measure. This could be explained by its substantial number of possible labels (174) and its relatively low label cardinality (26.04) which can often lead to wrong predictions being ranked as first. 

\subsection{Effect of the number of layers}

\begin{figure}[htpb]
\begin{subfigure}{0.98\textwidth}
    \centering
    \captionsetup{width=0.45\textwidth}
    \includegraphics[height = 0.7cm, width = 10.5cm]{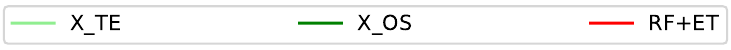}
\end{subfigure}

\begin{subfigure}{0.45\textwidth}
    \centering
    \captionsetup{width=0.45\textwidth}
    \includegraphics[width = 5cm,height=3.5cm]{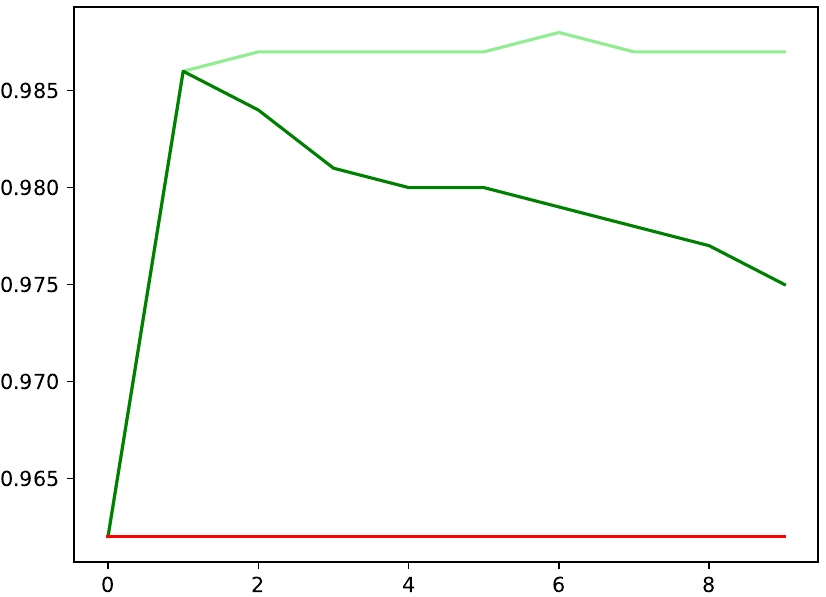}
    \caption{Genbase}
    \label{figure:genbasemicroAUC}
\end{subfigure}
\begin{subfigure}{0.45\textwidth}
    \centering
    \captionsetup{width=0.45\textwidth}
    \includegraphics[width = 5cm,height=3.5cm]{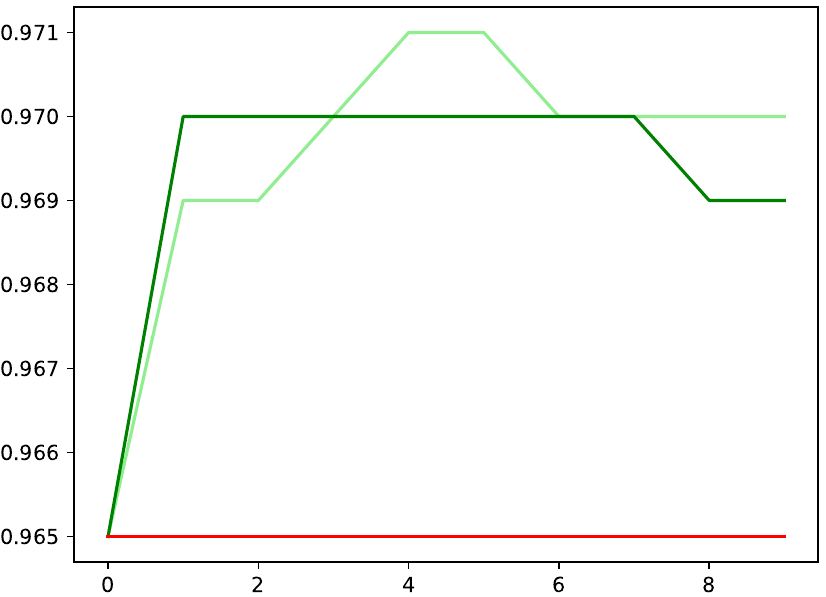}
    \caption{Medical}
    \label{figure:medicalmicroAUC}
\end{subfigure}

\begin{subfigure}{0.45\textwidth}
    \centering
    \captionsetup{width=0.45\textwidth}
    \includegraphics[width = 5cm,height=3.5cm]{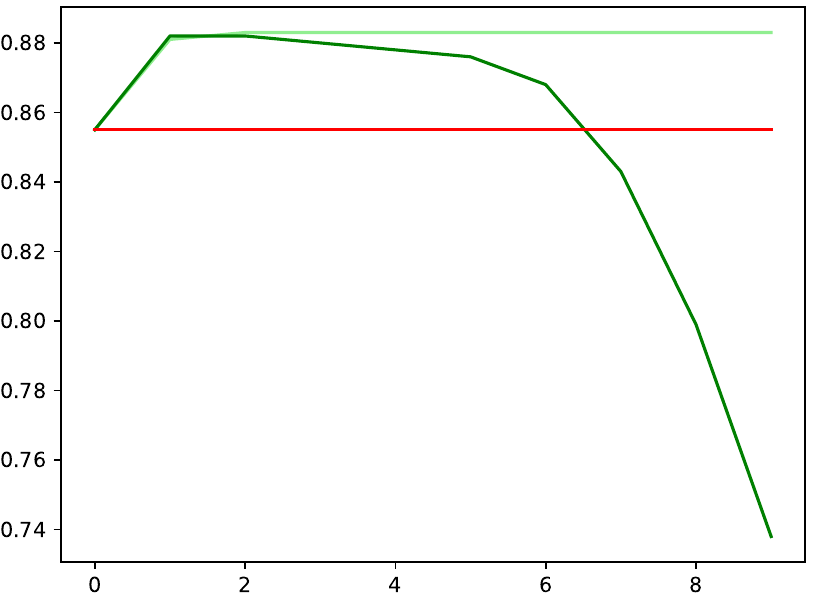}
    \caption{LLOG}
    \label{figure:LLOGmicroAUC}
\end{subfigure}
\begin{subfigure}{0.45\textwidth}
    \centering
    \captionsetup{width=0.45\textwidth}
    \includegraphics[width = 5cm,height=3.5cm]{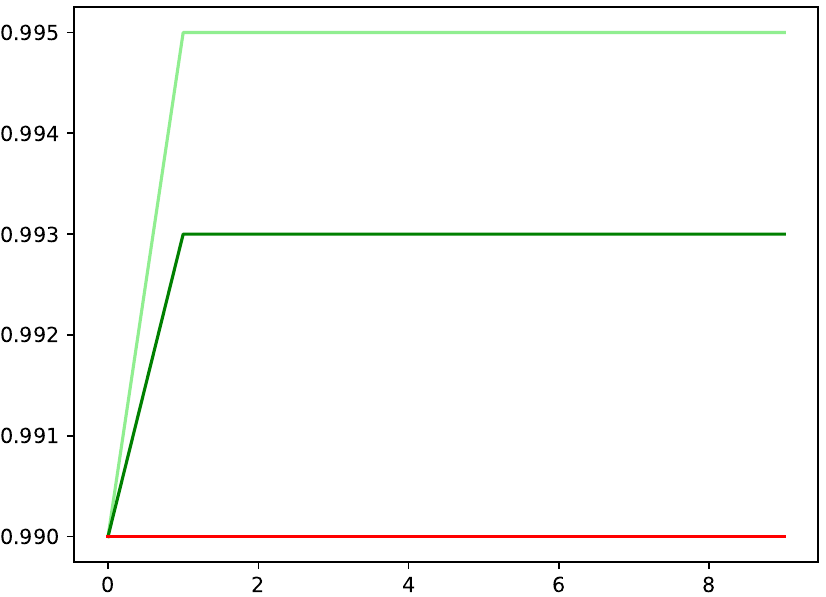}
    \caption{Gram\_negative}
    \label{figure:gramnegativemicroAUC}
\end{subfigure}
\caption{MicroAUC progression throughout the building of our model. The x-axis corresponds to the number of layers in the model, whereas the y-axis corresponds to the MicroAUC measured in the test partitions.}
\label{figure:microauclayers}
\end{figure}

In Figure \ref{figure:microauclayers}, we present the progression of microAUC, our optimized measure, throughout the training of our proposed model for some datasets. The values correspond to the mean value of 10 folds measured in the test partitions. 

As can be seen, the addition of the tree-embeddings is associated to an immediate improvement in the performance, attesting their representation power. Such increase in the performance can be followed by a plateau, where both training and testing performances present a static behaviour, as seen in Figure~\ref{figure:LLOGmicroAUC}. An oscillating trend was also noticed in some datasets, nonetheless such fluctuation tends to be minor.

The inclusion of output space (OS) features also contributes to the performance. However, its improvement is less pronounced. This difference is visible in Figure \ref{figure:gramnegativemicroAUC} where X\_TE approximates a flawless performance. In other situations, such as the ones depicted in Figures \ref{figure:genbasemicroAUC} and \ref{figure:LLOGmicroAUC}, X\_OS fails to hold a competitive performance, leading often to sub-optimal models.

\section{Conclusion}
\label{section:conclusion}

In this paper, we have proposed a novel deep tree-ensemble model for multi-output prediction tasks which can integrate tree-embeddings as well as output features in the learning process. Our results have shown that our model provides superior results in both multi-target regression and multi-label classification. 

More precisely, they reveal that integrating a representation learning component in each layer can significantly boost prediction performance, affirming our original hypothesis. Tree-embeddings can be more meaningful than just output features in both tasks. Additionally, in some cases, the combination of both can boost the performance even further. We also highlight that preserving the original feature set is essential to achieve high performance, since the tree-embeddings variant rarely yields superior results by itself. 

As for future work, we could investigate the incorporation of an extra steps in the prediction phase, such as the outlier removal performed by Moyano \cite{moyano2017}. In this case, models whose predictions deviate from the ensemble are not incorporated in the final prediction.

Moreover, this paper focuses on fundamental multi-output prediction. In the future, it would be very interesting to investigate the use of the proposed method in specialized application domains, such as the field of large scale image MLC, where our approach could potentially get combined with state of the art computer vision techniques. 

In addition, the investigation of the performance of our method in the field of large-scale data or big data would be a very interesting topic for future research. For example, we could investigate replacing PCA with a computationally more efficient feature transformation method.

Furthermore, we would like to exploit different ensembles of trees. That is, explore if the inclusion of other tree-based ensembles can improve the results, such as predictive bi-clustering trees \cite{pliakos2019} \cite{zamith2020}, extreme gradient boosting \cite{chen2016} and oblique forests \cite{katuwal2020}. Alternative representation components in the construction of the layers of our approach as well as deeper architectures for specific cases could be interesting research topics in the future. Finally, other multi-output tasks such as hierarchical multi-label classification \cite{nakano2019} and extreme multi-label classification \cite{bhatia2015} may also be addressed with our method, specially because the MLDF would result in a very high number of features whereas the tree-embeddings would be unaffected.

\section*{Acknowledgements}
The authors acknowledge the support from the Research Fund Flanders (through research project G080118N) and from the Flemish Government (AI Research Program).
\bibliography{main}

\end{document}